%% file: acl.tex
\pdfoutput=1

\documentclass[11pt]{article}

\usepackage[]{acl}

\usepackage{times}
\usepackage{latexsym}
\usepackage{multirow}
\usepackage{booktabs}
\usepackage{bm}
\usepackage{amsmath}
\usepackage{lscape}
\usepackage{CJKutf8}
\usepackage{booktabs}

\usepackage[T1]{fontenc}
\usepackage{devanagari}
\usepackage{graphicx}
\usepackage{makecell, threeparttable}

\usepackage{enumitem}

\usepackage[utf8]{inputenc}

\usepackage{microtype}

\title{Enhancing Cross-lingual Prompting with Dual Prompt Augmentation\Thanks{This work was supported by Alibaba Research Intern Program. It was done when Meng Zhou was an intern at Alibaba. Xin Li is the corresponding author.}}

\author{Meng Zhou$^{\heartsuit,\diamondsuit}$, Xin Li$^\diamondsuit$, Yue Jiang$^\heartsuit$, Lidong Bing$^\diamondsuit$ \\ {$^\heartsuit$Carnegie Mellon University} \\ {$^\diamondsuit$DAMO Academy, Alibaba Group} \\ {\texttt{\{mengzhou, yuejian2\}@andrew.cmu.edu}} \\ {\texttt{\{xinting.lx, l.bing\}@alibaba-inc.com}}}

\begin{document}
\maketitle
\begin{abstract}
Prompting shows promising results in few-shot scenarios. However, its strength for multilingual/cross-lingual problems has not been fully exploited. \citet{zhao-schutze-2021-discrete} made initial explorations in this direction by presenting that cross-lingual prompting outperforms cross-lingual finetuning. In this paper, we conduct an empirical exploration on the effect of each component in cross-lingual prompting and derive language-agnostic Universal Prompting, which helps alleviate the discrepancies between source-language training and target-language inference. Based on this, we propose DPA, a dual prompt augmentation framework, aiming at relieving the data scarcity issue in few-shot cross-lingual prompting. Notably, for XNLI, our method achieves 46.54\% with only 16 English training examples per class, significantly better than 34.99\% of finetuning. Our code is available at \url{https://github.com/DAMO-NLP-SG/DPA}.
\end{abstract}

\section{Introduction}


Although adapting Pre-trained Language Models (PLMs)~\citep{devlin-etal-2019-bert} to downstream NLP tasks via \textit{finetuning} is the de facto mainstream paradigm under fully supervised settings~\citep{wang-etal-2018-glue}, \textit{prompting}\footnote{In this work, the term ``prompting'' refers to prompt-based finetuning, where the parameters of PLMs are finetuned.}~\citep{gao-etal-2021-making,Radford2019LanguageMA,NEURIPS2020_1457c0d6,schick-schutze-2021-exploiting,schick-schutze-2021-just} has demonstrated its superiority over \textit{finetuning} in low-resource scenarios.
Typically, \textit{prompting} reformulates the classification task as a language modeling problem over manually-designed natural language prompts.

Despite the effectiveness of \textit{prompting} on English tasks, its potential for cross-lingual problems, which assume the availability of the training data in high-resource languages (e.g., \textit{English}) only, is still under-explored. \citet{zhao-schutze-2021-discrete} is the pioneering work to apply \textit{prompting} to cross-lingual NLP. However, their major efforts are spent on comparing different training strategies for cross-lingual prompting such as discrete prompting and soft prompting. They do not fully investigate the design choice of key components in \textit{prompting}, i.e., prompt template and verbalizer.

To provide a practical guide for designing cross-lingual prompting, we first conduct an empirical analysis to explore the effects of each \textit{prompting} component on the performance of cross-lingual transfer. 
Our preliminary study shows that template-free prompting combined with English-only inference, dubbed as language-agnostic ``Universal Prompting'' (UP) in this paper, generally performs well across different few-shot settings. 
Intuitively, UP avoids the discrepancies between the source-language training and the target-language inference, which intrinsically better fits cross-lingual tasks.

The derived UP is a concise solution with reasonable performance but does not take advantage of other available resources in the context of multilingual problems, e.g., the translation of verbalizers in target languages. Motivated by this fact, we propose a Dual Prompt Augmentation (DPA) framework to alleviate the data scarcity issue in few-shot scenarios. \textbf{Firstly}, 
we introduce multilingual verbalizers as answer augmentation for prompting, where the translated label tokens are treated as additional target-language supervision. 
\textbf{Secondly}, we propose prompt mixup as prompt input augmentation, which mixes the prompt representations in each batch. Intuitively, given two prompt representations on real data, we can generate a virtual representation based on their interpolation, which encodes the semantics in between. Our DPA framework is not task-dependent and does not require either external unlabeled data~\cite{xie2020unsupervised} or massive text manipulation efforts~\citep{wei-zou-2019-eda} compared with other data augmentation approaches.

\begin{table*}[!ht]
\resizebox{1.0\textwidth}{!}{%
  \centering\renewcommand{\arraystretch}{1.05} \setlength{\tabcolsep}{1.5pt}
\begin{tabular}{l|l|l|l}
 &    & Prompt Templates & Verbalizers           \\ \hline
\multicolumn{1}{c|}{\multirow{2}{*}{EN (source)}} & \citet{zhao-schutze-2021-discrete} & \underline{\smash{A}} \texttt{.} \texttt{Question:} \underline{\smash{B}} \texttt{?} \texttt{Answer:} <mask> \texttt{.} & Entailment: yes; Contradict: no; Neutral: maybe\\
\multicolumn{1}{c|}{}                    & Universal Prompting & \underline{\smash{A}} \texttt{.} \underline{\smash{B}} \texttt{?} <mask> \texttt{.} & Entailment: yes; Contradict: no; Neutral: maybe \\
\hline
\multicolumn{1}{c|}{\multirow{5}{*}{TR (target)}} & \citet{zhao-schutze-2021-discrete} & \underline{\smash{A}} \texttt{.} \texttt{Soru:} \underline{\smash{B}} \texttt{?} \texttt{Cevap:} <mask> \texttt{.} & Entailment: Evet; Contradict: hiçbir; Neutral: belki  \\
\multicolumn{1}{c|}{}                    & \text{ } \textsc{w/o Template Translation}  & \underline{\smash{A}} \texttt{.} \texttt{Question:}\underline{\smash{B}} \texttt{?} \texttt{Answer:} <mask> \texttt{.} & Entailment: Evet; Contradict: hiçbir; Neutral: belki \\
\multicolumn{1}{c|}{}                    & \text{ } \textsc{w/o Template Words}  & \underline{\smash{A}} \texttt{.} \underline{\smash{B}} \texttt{?} <mask> \texttt{.} & Entailment: Evet; Contradict: hiçbir; Neutral: belki \\
\multicolumn{1}{c|}{}                    & \text{ } \textsc{w/o Verbalizer Translation}  & \underline{\smash{A}} \texttt{.} \texttt{Soru:} \underline{\smash{B}} \texttt{?} \texttt{Cevap:} <mask> \texttt{.} & Entailment: yes; Contradict: no; Neutral: maybe \\
\multicolumn{1}{c|}{}                    & Universal Prompting  & \underline{\smash{A}} \texttt{.} \underline{\smash{B}} \texttt{?} <mask> \texttt{.} & Entailment: yes; Contradict: no; Neutral: maybe \\
\hline
\end{tabular}%
}
\caption{Prompt templates and verbalizers in English (EN) and
  Turkish (TR). A and B indicate two sentences of a sentence pair. For XNLI, A is the premise and B is the hypothesis. With the proposed language-agnostic Universal Prompting, we could treat source-language training and target-language inference in a unified fashion. }
\label{tab:languagepatterns}
\end{table*}

In summary, our contributions are as follows:
\begin{itemize}[leftmargin=3mm]
    \setlength\itemsep{-0.4em}
    \item We develop language-agnostic \textbf{Universal Prompting}, a concise prompting baseline with competitive performance for cross-lingual transfer. 
    \item To overcome the data scarcity issue, we propose \textbf{Dual Prompt Augmentation} for cross-lingual prompting to perform data augmentation from the views of prompt answers and prompt inputs. 
\end{itemize}

\section{Language-Agnostic Universal Prompting}
\label{sec:pilot}



In this section, we first empirically investigate the importance of essential elements, i.e., template and verbalizer design, in cross-lingual prompting~\citep{zhao-schutze-2021-discrete}. Based on our investigation, we derive a more competitive baseline called Universal Prompting. It is language-agnostic because it does not make assumptions about the input language in template design, and the verbalizer during training is taken for all other languages. Note that, since soft prompting (SP) and mixed prompting (MP) rely on an external bidirectional LSTM~\citep{Hochreiter1997LongSM} to create soft prompts and do not outperform discrete prompting (DP) significantly, we mainly discuss DP in this work for a clear comparison.

As illustrated in Table \ref{tab:languagepatterns}, 
\citet{zhao-schutze-2021-discrete} directly utilize the translated templates and verbalizers for target-language inference, making templates and verbalizers language-dependent. However, the translated templates are not seen and the translated verbalizers are never modeled by the PLM during training. This leads to discrepancies between the source-language training and the target-language inference.

To alleviate such discrepancies, we consider three possible variants.
 Specifically, these three variants are derived by avoiding translation on the template and verbalizer tokens or removing the template words, see 
Table~\ref{tab:languagepatterns} for concrete examples.





We follow the experimental setup  (refer to Section~\ref{sec:exp} for details) in \citet{zhao-schutze-2021-discrete} to evaluate the impact of the above designs \footnote{As we employ a different evaluation method, the reproduced results of \citet{zhao-schutze-2021-discrete} are slightly different from the original ones. More details can be found in Section~\ref{sec:exp}.}.
In Table~\ref{tab:pilot}, we observe that \textsc{w/o template translation} achieves slight but stable improvements under different shots. \textsc{w/o Template Words} simply removes the template words and achieves more obvious improvements. \textsc{w/o verbalizer translation} \footnote{Note that \textsc{w/o verbalizer translation} refers to not applying translated verbalizers during \textit{inference}. In Section~\ref{sec:method} we will show how to exploit the translated verbalizers as answer augmentation during \textit{training}.} avoids using translation at the verbalizer end and brings in the most significant improvements. Therefore, by alleviating discrepancies either in the aspect of verbalizer or template, the performance of cross-lingual prompting can be further improved.
By combining the advances of these variants, the Universal Prompting (UP) is derived to treat various languages in a unified fashion.
Specifically, UP alleviates the discrepancy of prompt templates and verbalizers simultaneously, which is a much stronger baseline than \citet{zhao-schutze-2021-discrete} in multilingual tasks. 



Note that the idea of removing template words in UP is distinct to ``null prompt''~\citep{LoganIV2021CuttingDO} from the perspective of motivation. ``Null prompt'' is proposed to simplify the manual prompt design on monolingual tasks. Compared with ``null prompt'', the primary goal of UP is to alleviate the source-target discrepancies in cross-lingual transfer. Moreover, besides removing template words, our UP also involves the design choice for target-language inference (\textsc{w/o Verbalizer Translation}), which proves to be a larger contribution according to the empirical results shown in Table~\ref{tab:pilot}. The effectiveness of using the verbalizer in the source language is also found in ~\cite{Lin2022FewshotLW}.

\begin{table}[t]
\centering \small
\begin{tabular}{|l|l|l|}
\hline
\textbf{Shots} & \textbf{Method} & \textbf{Accuracy} \\ 
\hline
\multirow{5}{*}{16} & \citet{zhao-schutze-2021-discrete} & $38.81_{1.61}$ \\
                    & \textsc{w/o Template Translation} & $39.15_{1.73}$ \\
                    & \textsc{w/o Template Words} & $39.87_{2.94}$ \\
                    & \textsc{w/o Verbalizer Translation} & $42.32_{1.81}$ \\
                    & Universal Prompting & $\bm{43.18}_{2.77}$ \\
\hline
\multirow{5}{*}{32} & \citet{zhao-schutze-2021-discrete} & $41.42_{1.66}$ \\
                    & \textsc{w/o Template Translation} & $41.72_{1.89}$ \\
                    & \textsc{w/o Template Words} & $43.66_{0.96}$ \\
                    & \textsc{w/o Verbalizer Translation} & $46.50_{1.54}$ \\
                    & Universal Prompting & $\bm{48.26}_{1.34}$ \\
\hline
\multirow{5}{*}{64} & \citet{zhao-schutze-2021-discrete} & $46.42_{0.65}$ \\
                    & \textsc{w/o Template Translation} & $46.75_{0.61}$ \\
                    & \textsc{w/o Template Words} & $47.60_{1.09}$ \\
                    & \textsc{w/o Verbalizer Translation} & $\bm{53.07}_{1.33}$ \\
                    & Universal Prompting & $52.19_{1.53}$ \\
\hline
\end{tabular}
\caption{Comparison results between \citet{zhao-schutze-2021-discrete} and its variants on XNLI. We calculate the average accuracy over 15 languages. The standard deviation over 5 runs is reported as the subscript.}
\label{tab:pilot}
\end{table}

\section{Dual Prompt Augmentation}
\label{sec:method}
In prompting, the mask token is directly used for making predictions. In this section, we formalize a Dual Prompt Augmentation (DPA) framework based on this crucial element of prompting.


\subsection{Prompt Answer Augmentation}
\label{sec:multi_verb}
In Section~\ref{sec:pilot}, we show that directly translating the verbalizers to the target language for inference is not helpful.
In this subsection, we explore the usage of verbalizer translation at the training stage.
Intuitively, their rich semantics could serve as high-quality paraphrases~\citep{Jiang2021HowCW} of the English verbalizer and provide additional supervision to train multilingual models. Motivated by this, we define a multilingual verbalizer for the English training data, which can be regarded as answer augmentation for the mask token. Formally, given the pre-built prompt $\bm{x}$ filled with input sentences, the training objective is to maximize the likelihood of verbalized label tokens in multiple languages:

\begin{equation}
\small
\label{eq:multi_verbalizer}
\begin{aligned}
\arg\max_{\bm{\theta}} \sum_{\bm{x}} \frac{1}{|\mathcal{L}|} \sum_{\bm{\ell}  \in \mathcal{L}} \log P\big(\left\langle \text{mask} \right\rangle = V_{\ell}(\bm{y}) | \bm{x};\bm{\theta}\big)
\end{aligned}
\end{equation}
where $\theta$ denotes the parameters of the PLM. $V_{\ell}$ is the verbalizer in a certain language $\ell \in \mathcal{L}$, and it maps from the gold label to a specific word in language $\ell$. \footnote{Please refer to Appx.~\ref{sec:add_imple} for the language set we use} In comparison, UP only takes $\mathcal{L} =$ \{EN\}, which is a monolingual verbalizer.

\subsection{Input Augmentation with Prompt Mixup}
Previous mixup methods for NLP perform the whole-sequence interpolation at the input embedding level~\citep{Zhang2021MixUpTL, Guo2019AugmentingDW} or hidden representation level~\citep{jindal-etal-2020-augmenting,chen-etal-2020-mixtext}. However, directly applying previous methods to prompting has been shown to even lead to a significant performance drop in ~\citet{Zhou2021FlipDAEA}. In prompting-based methods, the most important hidden space representation for classification is encoded at the position of mask tokens. Different training data may have different sequence lengths and their mask tokens are at different positions. The interpolation between the representation of a mask token and a normal verbal token would be meaningless in prompting. Therefore, we propose to interpolate between the top-most mask token representations to augment prompt inputs. Then the interpolated representation is fed into the masked language modeling head. 

Formally, let $\bm{m}_i = h(\bm{x}_{i})$ and $\bm{m}_j = h(\bm{x}_{j})$ be the top-most hidden representations corresponding to the mask tokens of two prompts $\bm{x}_i$ and $\bm{x}_j$, respectively. Then we perform linear interpolation to produce a virtual representation:
\begin{equation}
\label{eq:mixup_hidden}
\begin{aligned}
\hat{\bm{m}}_{ij} = \lambda h(\bm{x}_i) + (1 - \lambda) h(\bm{x}_j)
\end{aligned}
\end{equation}
where $\lambda$ follows a Beta distribution, i.e., $\lambda \sim \beta(\alpha, \alpha)$. The corresponding answer labels are linearly interpolated accordingly:
\begin{equation}
\label{eq:mixup_label}
\begin{aligned}
\hat{\bm{y}}_{ij} = \lambda \bm{y}_i + (1 - \lambda) \bm{y}_j
\end{aligned}
\end{equation}
Considering an augmented multilingual verbalizer as in Section~\ref{sec:multi_verb}, the training objective of this particular virtual example would be:
\begin{equation}
\small
\label{eq:mixup_loss}
\begin{aligned}
\arg\max_{\bm{\theta}} \frac{1}{|\mathcal{L}|} \sum_{\bm{\ell} \in \mathcal{L}} \big\{\lambda \log P\big(\left\langle \text{mask} \right\rangle = V_{\ell}(\bm{y}_i) | \hat{\bm{m}}_{ij};\bm{\theta}\big) \\ + (1 - \lambda) \log P\big(\left\langle \text{mask} \right\rangle = V_{\ell}(\bm{y}_j) | \bm{\hat{m}}_{ij};\bm{\theta}\big)\big\}
\end{aligned}
\end{equation}
The interpolation is performed in a dynamic in-batch fashion. For a mini-batch drawn from the training set, we will split it into pairs and generate a virtual prompt representation based on each pair.

\section{Experiments}
\label{sec:exp}
\subsection{Setup}
\paragraph{Datasets} We conduct experiments on two sentence-pair classification tasks: XNLI~\citep{conneau-etal-2018-xnli, williams-etal-2018-broad} for cross-lingual natural language inference and PAWS-X~\citep{yang-etal-2019-paws} for multilingual paraphrase identification. For these two datasets, while the evaluation data is human-translated, the golden training data is only available in English. 


\begin{table*}[ht]
\centering
\scriptsize
\setlength{\tabcolsep}{1.3mm}
\begin{tabular}{c|lcccccccccccccccc}
\toprule
 {\bf Shots} & {\bf Method} & {\bf EN} & {\bf AR} & {\bf BG} & {\bf DE} & {\bf EL} & {\bf ES} & {\bf FR} & {\bf HI} & {\bf RU} & {\bf SW} & {\bf TH} & {\bf TR} & {\bf UR} & {\bf VI} & {\bf ZH} & {\bf Avg.} \\
\midrule
\multirow{5}{*}{\textit{16}}
& FT & 35.62&35.11 &34.85 &35.07 &35.08 &35.21 &34.95 &34.89 &34.52 &35.07 &34.92 &34.79 &35.02 &35.02 &34.71 &34.99±1.84 \\ 
& PCT & 42.43 & 35.80 & 37.48 &36.02 &40.23 &36.14 &38.79 &39.79 &37.96 &36.32 &39.01 &37.41 &35.46 &38.84 &38.90 &38.04±3.52 \\ 
& UP & 47.68&42.01 &45.50 &44.51 &46.68 &36.61 &46.81 &40.29 &45.43 &42.06 &44.21 &41.04 &40.61 &45.79 &38.42 &43.18±2.77 \\
& DPA & 48.55 & \textbf{46.24} & \textbf{47.95} & \textbf{48.00} &47.41 & \textbf{47.47} & \textbf{48.61} & \textbf{44.36} &46.76 &44.35 &45.95 & \textbf{45.83} & \textbf{44.80} & \textbf{47.31} & \textbf{44.55} & \textbf{46.54±1.83} \\
& \text{ } \textsc{w/o MV} & \textbf{49.54} &41.55 &46.84 &45.53 &47.59 &34.63 &48.55 &42.39 & \textbf{47.18} &43.95 &\textbf{46.37} &43.82 &43.32 &46.52 &40.09 & 44.52±2.15 \\
& \text{ } \textsc{w/o Mixup} & 48.38 &45.59 &47.74 &47.72 & \textbf{47.60} &44.38 &47.83 &42.44 &46.69 &\textbf{44.38} &44.65 &45.52 &43.48 &46.65 &40.83 &45.59±1.91 \\
\midrule
\multirow{5}{*}{\textit{32}}
& FT & 37.62 & 36.82 & 36.61 & 37.03 & 37.07 & 37.39 & 37.53 & 37.35 & 36.83 & 36.42 & 36.40 & 36.40 & 36.71 & 36.84 & 36.96 & 36.93±1.96 \\ 
& PCT & 46.63 & 41.33 & 44.30 &43.35 &45.31 &45.61 &46.79 &43.32 &44.13 &40.88 &42.86 &43.19 &38.94 &44.85 &43.81 &43.69±2.11 \\ 
& UP & 53.33&47.70 &50.87 &49.74 &51.41 &41.48 &51.09 &44.97 &50.11 &46.76 &49.50 &45.92 &45.64 &51.00 &44.33 &48.26±1.34 \\
& DPA & 52.79 & \textbf{49.37} & \textbf{51.48} & \textbf{50.84} & \textbf{51.78} &50.05 &\textbf{51.77} & \textbf{48.08} & \textbf{50.46} &47.30 &49.35 &\textbf{50.14} &\textbf{47.44} &\textbf{50.84} & \textbf{48.25} & \textbf{49.99±2.21} \\
& \text{ } \textsc{w/o MV} & \textbf{53.75} &48.42 &50.71 &50.57 &51.76 &41.98 &51.54 &45.64 &\textbf{50.46} &45.84 &\textbf{49.65} &47.42 &45.58 &50.56 &47.54 &48.76±1.56 \\
& \text{ } \textsc{w/o Mixup} & 52.38 &49.29 &51.39 &50.76 &51.60 & \textbf{50.21} &51.54 &47.57 &50.35 &\textbf{47.56} &49.07 &49.56 &47.02 &50.65 &46.24 &49.68±1.46 \\
\midrule
\multirow{5}{*}{\textit{64}}
& FT & 42.97 &40.70 &41.29 &41.68 &42.09 &42.46 &42.23 &40.59 &40.38 &39.96 &40.65 &40.84 &40.24 &42.09 &40.53 &41.25±3.60 \\ 
& PCT & 52.26 & 46.39 & 48.73 &48.39 &49.64 &49.46 &50.46 &47.48 &48.52 &45.27 &48.28 &48.55 &44.76 &49.81 &49.12 &48.47±2.82 \\ 
& UP & 57.76&51.67 &54.85 &54.99 &54.69 &51.63 &54.96 &47.97 &53.32 &48.12 &51.91 &49.89 &47.86 &54.14 &49.13 &52.19±1.54 \\
& DPA & \textbf{59.97} &53.18 &56.51 & \textbf{56.67} &55.63 & \textbf{56.79} & \textbf{56.97} &51.77 &55.46 &50.71 &53.35 & \textbf{54.21} &50.76 &56.05 &53.09 &54.74±0.93 \\
& \text{ } \textsc{w/o MV} & 59.17 & \textbf{53.79} & \textbf{56.95} &56.53 & \textbf{56.18} &55.35 &56.48 & \textbf{52.17} & \textbf{55.72} & \textbf{50.89} & \textbf{54.55} &53.35 & \textbf{51.62} & \textbf{56.43} & \textbf{54.42} & \textbf{54.91±1.18} \\
& \text{ } \textsc{w/o Mixup} & 59.56 &53.06 &55.98 &55.65 &55.16 &56.67 &56.66 &51.44 &55.18 &49.99 &52.90 &53.76 &49.80 &55.43 &53.70 &54.33±0.98 \\
\midrule
\multirow{5}{*}{\textit{128}}
& FT & 47.24 &43.91 &44.13 &43.96 &44.38 &45.25 &44.48 &42.38 &42.81 &42.87 &42.87 &42.93 &42.36 &44.60 &42.87 &43.80±2.58 \\ 
& PCT & 55.31 & 48.55 & 52.09 &50.75 &52.92 &52.69 &52.79 &50.43 &51.60 &47.86 &50.88 &50.37 &48.04 &52.20 &51.79 &51.22±2.58 \\ 
& UP & 60.08&51.31 &56.60 &55.10 &56.17 &51.25 &56.97 &49.62 &55.18 &48.71 &53.87 &50.42 &49.20 &55.03 &53.15 &53.51±3.51 \\
& DPA & \textbf{62.57} & 54.91 & 58.72 & \textbf{58.81} & \textbf{58.25} &\textbf{59.47} &58.76 & \textbf{52.93} & \textbf{57.35} &50.95 &54.30 & \textbf{54.94} &51.47 &57.80 &54.99 & \textbf{56.42±1.37} \\
& \text{ } \textsc{w/o MV} & 61.51 & \textbf{55.31} &58.67 &58.15 &58.12 &58.10 &58.42 &52.31 &56.99 &50.80 & \textbf{55.40} &53.88 & \textbf{51.74} & \textbf{57.96} &\textbf{56.12} &56.23±0.90 \\
& \text{ } \textsc{w/o Mixup} & 61.84 &54.59 & \textbf{58.77} &58.57 &57.77 &59.13 & \textbf{58.89} &52.70 &56.99 & \textbf{52.05} &54.15 &54.69 &51.31 &57.27 &55.59 &56.29±1.46 \\
\midrule
\multirow{5}{*}{\textit{256}}
& FT & 59.49 &52.87 &55.92 &55.51 &55.07 &57.44 &56.32 &51.75 &54.19 &49.88 &52.38 &53.68 &50.38 &55.37 &53.95 &54.28±2.15 \\ 
& PCT & 60.09 & 53.51 & 57.21 &56.60 &57.63 &58.78 &58.42 &54.07 &56.35 &51.80 &54.57 &54.62 &50.56 &56.36 &56.14 &55.78±1.63 \\ 
& UP & 65.08&56.57 &61.03 &60.65 &60.74 &59.21 &61.01 &55.18 &59.41 &53.73 &57.66 &57.62 &54.08 &60.58 &58.71 &58.75±1.92 \\
& DPA & \textbf{67.97} & \textbf{59.54} & \textbf{63.59} & \textbf{63.26} &\textbf{62.34} &\textbf{64.80} &\textbf{63.93} &\textbf{58.39} &\textbf{61.87} &\textbf{55.83} &\textbf{59.19} &\textbf{60.32} &\textbf{56.00} &\textbf{62.41} &\textbf{61.29} &\textbf{61.38±0.92} \\
& \text{ } \textsc{w/o MV} & 65.80 &58.07 &62.04 &61.33 &61.05 &63.03 &62.36 &56.16 &60.14 &54.17 &58.23 &57.62 &54.12 &60.52 &59.81 &59.63±0.92 \\
& \text{ } \textsc{w/o Mixup} & 67.40 &58.02 &62.33 &62.18 &61.35 &63.61 &62.93 &56.89 &60.75 &54.68 &58.06 &59.00 &54.74 &61.17 &59.33 &60.16±0.97 \\
\bottomrule
\end{tabular}
\caption{Zero-shot cross-lingual transfer accuracy on XNLI. FT: finetuning; \textsc{MV}: Multilingual Verbalizer. Reported results are averaged with 5 random seeds.}
\label{table:xnli-results}
\end{table*}

\begin{table}[h]
\resizebox{0.46\textwidth}{!}{%
\centering
\scriptsize
\setlength{\tabcolsep}{1.3mm}
\begin{tabular}{c|lcccccccc}
\toprule
 {\bf Shots} & {\bf Method} & {\bf EN} & {\bf DE} & {\bf ES} & {\bf FR} & {\bf JA} & {\bf KO} & {\bf ZH} & {\bf Avg.} \\
\midrule
\multirow{5}{*}{\textit{256}}
& FT & 63.18&60.81 &60.95 &61.39 &58.60 &58.48 &59.78 &60.46±4.23 \\
& UP & 65.50 &62.21 &63.24 &62.82 &54.11 &54.30 &55.99 &59.74±4.12 \\
& DPA & \textbf{71.87} & \textbf{68.59} & \textbf{69.10} & \textbf{69.02} & \textbf{60.41} & \textbf{60.88} & \textbf{62.75} & \textbf{66.09±3.62} \\
& \text{ } \textsc{w/o MV} & 69.06 &66.26 &66.47 &65.79 &59.28 &58.34 &60.77 &63.71±4.37 \\
& \text{ } \textsc{w/o Mixup} & 70.95&67.14 &67.58 &67.63 &59.01 &60.44 &61.16 &64.84±2.91 \\
\midrule
\multirow{5}{*}{\textit{512}}
& FT & 77.64 &73.41 &73.19 &74.33 &65.55 &65.19 &68.25 &71.08±5.81  \\ 
& UP & 83.31&76.18 &77.63 &77.42 &63.41 &65.03 &68.06 &73.01±1.52 \\
& DPA & \textbf{84.97} & \textbf{78.63} &79.60 & \textbf{80.48} & \textbf{67.86} &68.13 &\textbf{72.34} &\textbf{76.00±1.04} \\
& \text{ } \textsc{w/o MV} & 84.81 &78.56 &\textbf{79.67} &79.64 &67.04 & \textbf{68.34} &71.50 &75.65±0.64 \\
& \text{ } \textsc{w/o Mixup} & 84.84 &77.85 &79.36 &79.69 &66.76 &68.03 &71.03 &75.37±2.00 \\
\bottomrule
\end{tabular}%
}
\caption{Zero-shot cross-lingual transfer accuracy on PAWS-X. FT:finetuning; \textsc{MV}: Multilingual Verbalizer. Reported results are averaged with 5 random seeds.}
\label{table:pawsx-results}
\end{table}

\paragraph{Evaluation} We conduct our experiments by training the XLM-R base model~\citep{conneau-etal-2020-unsupervised} on English. Then the model will be directly applied to other target languages, without using any training examples of the target language. To make a reasonable comparison between finetuning and prompting, we ensure finetuning to be better than a random guess on each language. Therefore, we randomly sample without replacement $K \in \{16, 32, 64, 128, 256\}$ per class for XNLI and $K \in \{256, 512\}$ per class for PAWS-X to construct the training set. Then we use the same number of shots from the validation split to select the best model~\citep{Perez2021TrueFL}. 

The evaluation of few-shot cross-lingual transfer can be with large variance and depend on data selection~\citep{zhang2021revisiting,zhao-etal-2021-closer,keung-etal-2020-dont}.
In our work, to faithfully reflect the few-shot performance, separate training/validation sets are sampled for different runs.



\subsection{Results}
\paragraph{UP v.s. Finetuning/PCT} 
On the XNLI dataset, even the simplest prompting method for cross-lingual transfer, namely UP, consistently outperforms the finetuning (FT) method by a large margin. Besides, our language-agnostic UP also surpasses FT on the majority of languages on the more challenging PAWS-X. These observations suggest that prompting is indeed a better solution for few-shot learning in cross-language scenarios and our UP can serve as a strong baseline for cross-lingual prompting. We also reproduce PCT~\citep{qi-etal-2022-enhancing}, another recent cross-lingual prompting method based on data augmentation and consistency training, with our evaluation method. Table~\ref{table:xnli-results} shows that UP outperforms PCT consistently without any data augmentation approach or introducing additional loss terms.

\paragraph{Dual Prompt Augmentation} 

With the proposed DPA framework, our prompting method achieves consistent improvement over UP, indicating that multilingual verbalizers from the answer view and prompt mixup from the input view are both effective ways to enhance cross-lingual prompting. The comparison results in Table~\ref{table:xnli-results} and Table~\ref{table:pawsx-results} also exhibit clear superiority of our method over cross-lingual finetuning. Even in the most resource-rich settings, compared to FT, our method still obtains 7.1\% (256 shots) and 4.9\% (512 shots) absolute gains on XNLI and PAWS-X.


\paragraph{Ablation Study} The performance of our prompting method will become worse when removing either prompt mixup or multilingual verbalizer, showing that both prompt input and prompt answer augmentation contribute positively to the improvement. We also notice that the negative effects brought by \textsc{DPA w/o MV} are generally larger, showing the necessity of target-language guidance for cross-lingual prompting.

\subsection{Inference Strategy}

\label{sec:infer_strategy}
A natural extension for the DPA framework is to leverage the multilingual verbalizer in some way for target-language inference as well. For comparisons, we heuristically devise the following inference strategies :
\paragraph{(1) English Verbalizer} The English verbalizer is still used when transferring to target languages. This strategy is used to produce results in Table~\ref{table:xnli-results} and ~\ref{table:pawsx-results}. To formalize:
\begin{equation}
\label{eq:EN_verbalizer}
\begin{aligned}
\hat{\bm{y}} = \arg\max_{\bm{y}} P\big(\left\langle \text{mask} \right\rangle = V_{EN}(\bm{y}) | \bm{x};\bm{\theta}\big)
\end{aligned}
\end{equation}
\paragraph{(2) Target Language Verbalizer} The verbalizer in the corresponding target language is used, which is the practice of \citet{zhao-schutze-2021-discrete} during inference time. However, in this case, our DPA framework has already modeled these words during the training time. To formalize:
\begin{equation}
\label{eq:target_verbalizer}
\begin{aligned}
\hat{\bm{y}} = \arg\max_{\bm{y}} P\big(\left\langle \text{mask} \right\rangle = V_{target}(\bm{y}) | \bm{x};\bm{\theta}\big)
\end{aligned}
\end{equation}
\paragraph{(3) Taking Maximum over the Multilingual Verbalizer} In this strategy, we will take the maximum probability over the whole multilingual verbalizer. To formalize:
\begin{equation}
\label{eq:maximum_verbalizer}
\begin{aligned}
\hat{\bm{y}} = \arg\max_{\bm{y}, \bm{\ell}} P\big(\left\langle \text{mask} \right\rangle = V_{\ell}(\bm{y}) | \bm{x};\bm{\theta}\big)
\end{aligned}
\end{equation}

\paragraph{(4) Taking Sum over the Multilingual Verbalizer} In this strategy, we will take the sum of probability over the whole multilingual verbalizer. To formalize:
\begin{equation}
\label{eq:sum_verbalizer}
\begin{aligned}
\hat{\bm{y}} = \arg\max_{\bm{y}} \sum_{\bm{\ell} \in \mathcal{L}} P\big(\left\langle \text{mask} \right\rangle = V_{\ell}(\bm{y}) | \bm{x};\bm{\theta}\big)
\end{aligned}
\end{equation}

\paragraph{(5) Bilingual Verbalizer} In this strategy, we will take the sum of probability over the target language verbalizer and the English verbalizer. To formalize, the predicted label $\hat{y}$ is given by:
\begin{equation}
\label{eq:bilingual_verbalizer}
\begin{aligned}
\hat{\bm{y}} = \arg\max_{\bm{y}}\{P\big(\left\langle \text{mask} \right\rangle = V_{EN}(\bm{y}) | \bm{x};\bm{\theta}\big) \\+ P\big(\left\langle \text{mask} \right\rangle = V_{target}(\bm{y}) | \bm{x};\bm{\theta}\big) \}
\end{aligned}
\end{equation}

We use the checkpoint of XLM-R trained by 128 shots on the XNLI dataset and make inference with different strategies. Table~\ref{tab:infer_strategy} shows the accuracy by employing different inference strategies. We show that with our DPA framework, the inference is quite robust to the utilization of the verbalizer. This can probably be attributed to answer augmentation via multilingual verbalizers, which help to model label tokens in multiple languages. We choose to simply employ English-only inference due to its simplicity and slightly better performance to produce results in Tables~\ref{table:xnli-results} and ~\ref{table:pawsx-results}.

\begin{table}[!t]
  \centering \small
  
  \begin{tabular}{cc}
    \toprule
    Strategy Num. & Accuracy \\
    \midrule
    1 & \bm{$56.42_{1.37}$} \\
    2 & $56.31_{1.15}$ \\
    3 & $56.23_{1.09}$ \\
    4 & $56.33_{1.11}$ \\
    5 & $56.39_{1.21}$ \\
    \bottomrule
  \end{tabular}
  \caption{Accuracy of different inference strategies, averaged over 15 testing languages of XNLI and 5 random seeds.}
  \label{tab:infer_strategy}
\end{table}

\section{Conclusion} 
In this paper, we first derive language-agnostic Universal Prompting, a concise but competitive baseline for cross-lingual prompting. The proposed DPA framework can further enhance cross-lingual prompting as shown on two sentence-pair classification tasks. In the future, we will consider verifying the effectiveness of prompting and the DPA framework in cross-lingual sequence tagging or question-answering tasks~\cite{xu2022mpmr}.

\section{Limitations}
Our work mainly focuses on cross-lingual sentence-pair classification tasks. While it is directly applicable to single-sentence classification tasks~\cite{li-etal-2020-sunsupervised, ye-etal-2020-feature} but may require additional efforts to adapt our DPA framework to more complex cross-lingual tasks such as sequence tagging~\cite{liu-etal-2021-mulda, zhou-etal-2022-conner, zhou-etal-2023-improving, zhang-etal-2021-cross} or question answering~\cite{xu2022clozing, xu2022mpmr}. Another limitation is that the proposed multilingual verbalizer in the DPA framework requires an external machine translator to produce the translated verbalizers. Finally, we limit the language set of the multilingual verbalizer to the set of target languages in a multilingual dataset. Extending this language set might give us greater improvement for cross-lingual tasks.

\bibliography{anthology,custom}
\bibliographystyle{acl_natbib}

\newpage
\appendix
\clearpage


\section{Additional Implementation Details}
\label{sec:add_imple}
\paragraph{Implementation Package} Our implementation is based on PyTorch~\citep{Paszke2019PyTorchAI} and Huggingface Transformer~\citep{Wolf2019HuggingFacesTS} framework.
\paragraph{Model Details}
XLM-R base model, containing 270M parameters, is pretrained on 2.5TB of filtered CommonCrawl on 100 languages. It contains 12 Transformer layers with hidden space dimensions of 768 and 12 attention heads in each layer.
\paragraph{Computing Infrastructure}
All of our experiments are conducted on a single \textit{Tesla V100-SXM2 32G}. Gradient accumulation steps of 4 is used for prompting to overcome resource limitations. 
\paragraph{Hyperparameter Settings} Our major hyperparameter settings follow \citet{zhao-schutze-2021-discrete}. A fixed learning rate (1e-5) is used for all of our experiments without any learning rate schedule to compare finetuning with prompting~\citep{le-scao-rush-2021-many}. We use a smaller batch size of 8 for finetuning and prompting because it achieves slightly better performance. We use the max sequence length of 256. The model is trained for 50 epochs and we select the checkpoint by validation accuracy for testing as suggested in ~\citet{mosbach2021on, zhang2021revisiting}. The $\alpha$ value for $\beta$ distribution in prompt mixup is set to 1.2 for all of the experiments. 

\paragraph{Prompting}
The language sets $\mathcal{L}$ used for multilingual verbalizers are determined by the language availability of the dataset. Specifically, for XNLI, $\mathcal{L} = $ \{EN, AR, BG, DE, EL, ES, FR, HI, RU, SW, TH, TR, UR, VI, ZH\}. For PAWS-X, $\mathcal{L} = $ \{EN, DE, ES, FR, JA, KO, ZH\}

For simplicity, the verbalizers of target languages are translated by Google Translate. Similar with XNLI, we use "paraphrase $\rightarrow$ yes" and "non-paraphrase $\rightarrow$ no" as the verbalizer of PAWS-X in English. Table~\ref{tab:verbalizers} presents the full multilingual verbalizer we use for the PAWS-X dataset.

We discuss Universal Prompting across languages for multilingual sentence-pair classification tasks in Section~\ref{sec:pilot}. Moreover, we believe the same notion of alleviating source-target discrepancies in terms of prompt template and verbalizer is generally applicable for cross-lingual tasks, which is left for future work.

\begin{table}[t]
  \small\centering \setlength{\tabcolsep}{3pt}
\begin{tabular}{l|l}
                                        Language & Verbalizer           \\ \hline
\multicolumn{1}{c|}{\multirow{2}{*}{EN}}  & Paraphrase $\rightarrow$ yes \\
\multicolumn{1}{c|}{}                    & Non-paraphrase $\rightarrow$  no  \\ \hline
\multicolumn{1}{c|}{\multirow{2}{*}{DE}}  & Paraphrase $\rightarrow$ \ Ja \\
\multicolumn{1}{c|}{}                    & Non-paraphrase $\rightarrow$ Nein \\\hline
\multicolumn{1}{c|}{\multirow{2}{*}{ES}}  & Paraphrase $\rightarrow$ sí \\
\multicolumn{1}{c|}{}                    & Non-paraphrase $\rightarrow$ no\\ \hline
\multicolumn{1}{c|}{\multirow{2}{*}{FR}}  & Paraphrase $\rightarrow$ Oui \\
\multicolumn{1}{c|}{}                    & Non-paraphrase $\rightarrow$  non  \\ \hline
\multicolumn{1}{c|}{\multirow{2}{*}{JA}} & Paraphrase $\rightarrow$ \begin{CJK*}{UTF8}{gbsn}はい\end{CJK*} \\
\multicolumn{1}{c|}{}                    & Non-paraphrase $\rightarrow$ \begin{CJK*}{UTF8}{gbsn}ない\end{CJK*} \\\hline
\multicolumn{1}{c|}{\multirow{2}{*}{ZH}} & Paraphrase $\rightarrow$ \begin{CJK*}{UTF8}{gbsn}是\end{CJK*} \\
\multicolumn{1}{c|}{}                    & Non-paraphrase $\rightarrow$  \begin{CJK*}{UTF8}{gbsn}否\end{CJK*}  \\ \hline
\multicolumn{1}{c|}{\multirow{2}{*}{KO}} & Paraphrase $\rightarrow$ \begin{CJK*}{UTF8}{mj}예\end{CJK*} \\
\multicolumn{1}{c|}{}                    & Non-paraphrase $\rightarrow$  \begin{CJK*}{UTF8}{mj}아니\end{CJK*} \\\hline

\end{tabular}
\caption{The multilingual verbalizer for PAWS-X.}
\label{tab:verbalizers}
\end{table}

\section{Generalizability of Prompting Word Removal}
In Section~\ref{sec:pilot}, we show that by removing template words, UP provides a more reasonable baseline for cross-lingual prompting on XNLI. To see whether such a removal generalizes to other cross-lingual sentence-pair classification task, we also investigate the impact of removing template words on PAWS-X, as shown in Table~\ref{tab:generalization}. We find that UP still performs reasonably well on PAWS-X without template words. It was also shown in \citet{LoganIV2021CuttingDO} that hand-engineering prompt is less important when PLMs are finetuned for monolingual tasks. Our UP generalizes this in cross-lingual tasks.


\begin{table}[t]
\centering \small
\begin{tabular}{|l|l|l|}
\hline
\textbf{Shots} & \textbf{Method} & \textbf{Accuracy} \\ 
\hline
\multirow{2}{*}{256} & Universal Prompting & $59.74_{4.12}$ \\
                     & \textsc{w/ Template Words} & $57.01_{2.64}$ \\
\hline
\multirow{2}{*}{512} & Universal Prompting & $73.01_{1.52}$ \\
                     & \textsc{w/ Template Words} & $73.39_{2.54}$ \\
\hline
\end{tabular}
\caption{The ablation study of the impact of removing template words on PAWS-X. We calculate the average accuracy over 7 languages. The standard deviation over 5 runs is reported as the subscript.}
\label{tab:generalization}
\end{table}

\section{Performance with Standard Deviation}
In Table~\ref{table:xnli-variance} and ~\ref{table:pawsx-variance}, we show the performance with standard deviation specifically in every language.
\clearpage
\begin{landscape}
\input{xnli-variance}
\clearpage
\input{paws-x-variance}
\end{landscape}

\end{document}

%% file: xnli-variance.tex
\begin{table}[t]
\centering
\scriptsize
\setlength{\tabcolsep}{3pt}
\renewcommand{\arraystretch}{1.1}
\begin{tabular}{c|lcccccccccccccccc}
\toprule
 & {\bf Method} & {\bf EN} & {\bf AR} & {\bf BG} & {\bf DE} & {\bf EL} & {\bf ES} & {\bf FR} & {\bf HI} & {\bf RU} & {\bf SW} & {\bf TH} & {\bf TR} & {\bf UR} & {\bf VI} & {\bf ZH} & {\bf Avg.} \\
\midrule
\multirow{5}{*}{\textit{16shots}}
& FT & $35.62_{2.45}$ & $35.11_{1.53}$ & $34.85_{1.87}$ &$35.07_{2.11}$ &$35.08_{2.22}$ &$35.21_{2.20}$ & $34.95_{2.09}$ & $34.89_{1.84}$ & $34.52_{1.56}$ & $35.07_{2.01}$ & $34.92_{1.50}$ & $34.79_{1.77}$ & $35.02_{1.70}$ & $35.02_{1.89}$ & $34.71_{1.52}$ & $34.99_{1.84}$ \\ 
& UP & $47.68_{1.65}$ & $42.01_{3.68}$ & $45.50_{2.32}$ & $44.51_{2.43}$ & $46.68_{2.75}$ & $36.61_{4.18}$ & $46.81_{1.85}$ & $40.29_{5.19}$ & $45.43_{2.11}$ & $42.06_{4.00}$ & $44.21_{3.64}$ & $41.04_{4.19}$ & $40.61_{4.76}$ & $45.79_{2.10}$ & $38.42_{5.54}$ & $43.18_{2.77}$ \\
& \textsc{Ours} & $48.55_{1.43}$ & $\textbf{46.24}_{2.61}$ & $\textbf{47.95}_{1.72}$ & $\textbf{48.00}_{1.53}$ & $47.41_{2.27}$ & $\textbf{47.47}_{1.83}$ & $\textbf{48.61}_{1.48}$ & $\textbf{44.36}_{2.48}$ &$46.76_{1.55}$ &$44.35_{1.88}$ &$45.95_{2.41}$ & $\textbf{45.83}_{1.51}$ & $\textbf{44.80}_{2.20}$ & $\textbf{47.31}_{1.67}$ & $\textbf{44.55}_{3.76}$ & $\textbf{46.54}_{1.83}$ \\
& \text{ } \textsc{w/o MV} & $\textbf{49.54}_{2.62}$ &$41.55_{3.56}$ &$46.84_{2.05}$ &$45.53_{2.39}$ &$47.59_{2.59}$ &$34.63_{1.11}$ & $48.55_{2.21}$ & $42.39_{4.25}$ & $\textbf{47.18}_{1.87}$ &$43.95_{2.40}$ & $\textbf{46.37}_{2.31}$ & $43.82_{3.92}$ &$43.32_{3.66}$ & $46.52_{2.31}$ &$40.09_{3.51}$ & $44.52_{2.15}$ \\
& \text{ } \textsc{w/o Mixup} & $48.38_{1.80}$ & $45.59_{1.96}$ & $47.74_{1.94}$ & $47.72_{2.07}$ & $\textbf{47.60}_{2.57}$ & $44.38_{5.78}$ &$47.83_{1.54}$ & $42.44_{3.31}$ & $46.69_{1.31}$ & $\textbf{44.38}_{1.39}$ &$44.65_{2.53}$ & $45.52_{2.02}$ &$43.48_{2.66}$ &$46.65_{1.67}$ & $40.83_{4.31}$ &$45.59_{1.91}$ \\
\midrule
\multirow{5}{*}{\textit{32shots}}
& FT & $37.62_{2.66}$ & $36.82_{2.09}$ & $36.61_{2.28}$ & $37.03_{2.56}$ & $37.07_{1.90}$ & $37.39_{2.17}$ & $37.53_{1.71}$ & $37.35_{1.55}$ & $36.83_{2.37}$ & $36.42_{1.93}$& $36.40_{1.72}$ & $36.40_{1.59}$ & $36.71_{1.50}$ & $36.84_{2.32}$ & $36.96_{2.17}$ & $36.93_{1.96}$ \\ 
& UP & $53.33_{1.60}$& $47.70_{3.58}$ &$50.87_{1.41}$ &$49.74_{2.66}$ &$51.41_{1.98}$ &$41.48_{4.77}$ &$51.09_{1.49}$ &$44.97_{1.40}$ &$50.11_{0.76}$ &$46.76_{3.00}$ & $49.50_{1.36}$ & $45.92_{2.53}$ & $45.64_{2.07}$ &$51.00_{1.54}$ &$44.33_{2.51}$ &$48.26_{1.34}$ \\
& \textsc{Ours} & $52.79_{2.14}$ & $\textbf{49.37}_{2.85}$ & $\textbf{51.48}_{1.79}$ & $\textbf{50.84}_{1.98}$ & $\textbf{51.78}_{2.07}$ &$50.05_{3.60}$ & $\textbf{51.77}_{2.13}$ & $\textbf{48.08}_{1.84}$ & $\textbf{50.46}_{1.89}$ &$47.30_{2.89}$ &$49.35_{2.30}$ & $\textbf{50.14}_{1.77}$ &$\textbf{47.44}_{2.30}$ &$\textbf{50.84}_{2.50}$ & $\textbf{48.25}_{4.08}$ & $\textbf{49.99}_{2.21}$ \\
& \text{ } \textsc{w/o MV} & $\textbf{53.75}_{1.53}$ & $48.42_{1.06}$ & $50.71_{1.67}$ & $50.57_{1.21}$ & $51.76_{1.76}$ & $41.98_{4.85}$ & $51.54_{1.78}$ & $45.64_{3.76}$ & $\textbf{50.46}_{1.14}$ & $45.84_{3.32}$ & $\textbf{49.65}_{0.91}$ & $47.42_{3.57}$ & $45.58_{2.57}$ & $50.56_{1.63}$ & $47.54_{2.30}$ & $48.76_{1.56}$ \\
& \text{ } \textsc{w/o Mixup} & $52.38_{2.50}$ &$49.29_{2.05}$ &$51.39_{1.71}$ & $50.76_{1.75}$ & $51.60_{1.44}$ & $\textbf{50.21}_{3.00}$ & $51.54_{1.83}$ & $47.57_{0.51}$ & $50.35_{1.66}$ & $\textbf{47.56}_{1.02}$ &$49.07_{1.76}$ & $49.56_{1.25}$ & $47.02_{1.23}$ & $50.65_{1.45}$ & $46.24_{2.07}$ & $49.68_{1.46}$ \\
\midrule
\multirow{5}{*}{\textit{64shots}}
& FT & $42.97_{3.88}$ & $40.70_{4.06}$ & $41.29_{3.93}$ & $41.68_{3.66}$ &$42.09_{3.81}$ & $42.46_{4.09}$ & $42.23_{4.38}$ & $40.59_{3.39}$ & $40.38_{3.58}$ & $39.96_{3.10}$ & $40.65_{3.48}$ & $40.84_{3.26}$ & $40.24_{3.01}$ & $42.09_{3.68}$ & $40.53_{3.46}$ & $41.25_{3.60}$ \\ 
& UP & $57.76_{1.49}$ & $51.67_{2.31}$ & $54.85_{1.52}$ & $54.99_{1.71}$ & $54.69_{1.00}$ & $51.63_{3.98}$ & $54.96_{0.87}$ & $47.97_{1.74}$ & $53.32_{1.73}$ & $48.12_{1.21}$ & $51.91_{1.27}$ & $49.89_{2.80}$ & $47.86_{2.10}$ & $54.14_{1.21}$ & $49.13_{3.38}$ & $52.19_{1.54}$ \\
& \textsc{Ours} & $\textbf{59.97}_{1.25}$ & $53.18_{1.12}$ & $56.51_{1.14}$ & $\textbf{56.67}_{1.12}$ & $55.63_{0.68}$ & $\textbf{56.79}_{1.17}$ & $\textbf{56.97}_{1.52}$ & $51.77_{1.09}$ & $55.46_{1.26}$ & $50.71_{1.50}$ &$53.35_{1.36}$ & $\textbf{54.21}_{1.06}$ &$50.76_{1.39}$ &$56.05_{0.97}$ &$53.09_{1.20}$ &$54.74_{0.93}$ \\
& \text{ } \textsc{w/o MV} & $59.17_{1.59}$ & $\textbf{53.79}_{1.66}$ & $\textbf{56.95}_{0.90}$ & $56.53_{1.23}$ & $\textbf{56.18}_{0.99}$ &$55.35_{3.39}$ &$56.48_{1.96}$ & $\textbf{52.17}_{1.22}$ & $\textbf{55.72}_{1.18}$ & $\textbf{50.89}_{1.36}$ & $\textbf{54.55}_{0.91}$ &$53.35_{1.27}$ & $\textbf{51.62}_{1.05}$ & $\textbf{56.43}_{1.14}$ & $\textbf{54.42}_{1.04}$ & $\textbf{54.91}_{1.18}$ \\
& \text{ } \textsc{w/o Mixup} & $59.56_{1.22}$ &$53.06_{1.02}$ &$55.98_{1.05}$ & $55.65_{1.01}$ & $55.16_{0.48}$ & $56.67_{1.10}$ & $56.66_{1.30}$ &$51.44_{1.45}$ &$55.18_{1.18}$ &$49.99_{1.60}$ &$52.90_{0.85}$ &$53.76_{0.92}$ &$49.80_{1.89}$ &$55.43_{0.87}$ &$53.70_{1.45}$ & $54.33_{0.98}$ \\
\midrule
\multirow{5}{*}{\textit{128shots}}
& FT & $47.24_{4.50}$ & $43.91_{2.43}$ & $44.13_{2.63}$ & $43.96_{2.83}$ &$44.38_{2.13}$ & $45.25_{3.38}$ & $44.48_{2.89}$ & $42.38_{2.68}$ & $42.81_{2.42}$ & $42.87_{1.77}$ & $42.87_{2.48}$ & $42.93_{2.53}$ &$42.36_{2.37}$ & $44.60_{2.74}$ & $42.87_{2.57}$ & $43.80_{2.58}$ \\ 
& UP & $60.08_{2.56}$& $51.31_{6.02}$ & $56.60_{3.30}$ & $55.10_{3.84}$ &$56.17_{1.81}$ & $51.25_{10.31}$ & $56.97_{2.19}$ & $49.62_{4.33}$ & $55.18_{1.75}$ & $48.71_{3.94}$ & $53.87_{2.05}$ & $50.42_{3.79}$ &$49.20_{3.39}$ & $55.03_{4.04}$ & $53.15_{2.61}$ & $53.51_{3.51}$ \\
& \textsc{Ours} & $\textbf{62.57}_{1.69}$ & $54.91_{1.93}$ & $58.72_{1.31}$ & $\textbf{58.81}_{1.49}$ & $\textbf{58.25}_{0.99}$ & $\textbf{59.47}_{1.79}$ & $58.76_{1.26}$ & $\textbf{52.93}_{1.65}$ & $\textbf{57.35}_{1.38}$ & $50.95_{1.34}$ & $54.30_{2.39}$ & $\textbf{54.94}_{1.76}$ & $51.47_{2.37}$ & $57.80_{1.50}$ & $54.99_{1.86}$ & $\textbf{56.42}_{1.37}$ \\
& \text{ } \textsc{w/o MV} & $61.51_{1.58}$ & $\textbf{55.31}_{1.26}$ & $58.67_{1.38}$ & $58.15_{1.83}$ & $58.12_{1.02}$ & $58.10_{1.74}$ & $58.42_{0.50}$ & $52.31_{1.03}$ & $56.99_{1.09}$ & $50.80_{1.30}$ & $\textbf{55.40}_{0.97}$ & $53.88_{0.81}$ & $\textbf{51.74}_{0.76}$ & $\textbf{57.96}_{1.32}$ & $\textbf{56.12}_{1.50}$ & $56.23_{0.90}$ \\
& \text{ } \textsc{w/o Mixup} & $61.84_{1.64}$ & $54.59_{1.41}$ & $\textbf{58.77}_{1.55}$ & $58.57_{1.24}$ & $57.77_{1.74}$ & $59.13_{1.92}$ & $\textbf{58.89}_{1.34}$ & $52.70_{1.86}$ & $56.99_{1.83}$ & $\textbf{52.05}_{1.71}$ & $54.15_{1.58}$ & $54.69_{1.38}$ & $51.31_{1.07}$ &$57.27_{1.92}$ & $55.59_{2.44}$ & $56.29_{1.46}$ \\
\midrule
\multirow{5}{*}{\textit{256shots}}
& FT & $59.49_{2.11}$ & $52.87_{2.34}$ & $55.92_{2.15}$ & $55.51_{2.25}$ & $55.07_{2.60}$ & $57.44_{1.74}$ & $56.32_{2.04}$ & $51.75_{2.72}$ & $54.19_{2.46}$ & $49.88_{2.51}$ & $52.38_{2.96}$ & $53.68_{1.80}$ & $50.38_{2.34}$ & $55.37_{2.02}$ & $53.95_{2.08}$ & $54.28_{2.15}$ \\ 
& UP & $65.08_{1.49}$ & $56.57_{2.08}$ & $61.03_{1.86}$ & $60.65_{1.27}$ & $60.74_{1.84}$ & $59.21_{2.64}$ & $61.01_{2.41}$ & $55.18_{2.54}$ & $59.41_{2.29}$ & $53.73_{2.27}$ & $57.66_{2.08}$ & $57.62_{2.65}$ & $54.08_{2.47}$ & $60.58_{1.84}$ & $58.71_{2.16}$ & $58.75_{1.92}$ \\
& \textsc{Ours} & $\textbf{67.97}_{1.02}$ & $\textbf{59.54}_{1.36}$ & $\textbf{63.59}_{1.23}$ & $\textbf{63.26}_{1.23}$ & $\textbf{62.34}_{0.91}$ &$\textbf{64.80}_{1.17}$ & $\textbf{63.93}_{1.06}$ & $\textbf{58.39}_{0.71}$ & $\textbf{61.87}_{1.55}$ & $\textbf{55.83}_{1.52}$ & $\textbf{59.19}_{1.06}$ &$\textbf{60.32}_{1.08}$ & $\textbf{56.00}_{0.95}$ & $\textbf{62.41}_{0.52}$ & $\textbf{61.29}_{1.14}$ & $\textbf{61.38}_{0.92}$ \\
& \text{ } \textsc{w/o MV} & $65.80_{1.25}$ & $58.07_{1.22}$ & $62.04_{1.60}$ & $61.33_{1.45}$ & $61.05_{1.24}$ & $63.03_{1.43}$ & $62.36_{1.22}$ & $56.16_{0.52}$ & $60.14_{1.53}$ & $54.17_{1.00}$ & $58.23_{1.24}$ & $57.62_{1.02}$ & $54.12_{0.95}$ & $60.52_{1.34}$ & $59.81_{1.24}$ & $59.63_{0.92}$ \\
& \text{ } \textsc{w/o Mixup} & $67.40_{0.65}$ & $58.02_{1.41}$ & $62.33_{1.47}$ & $62.18_{1.00}$ & $61.35_{1.24}$ & $63.61_{1.24}$ & $62.93_{1.25}$ & $56.89_{1.06}$ & $60.75_{1.67}$ & $54.68_{1.60}$ & $58.06_{1.51}$ & $59.00_{1.44}$ & $54.74_{1.10}$ & $61.17_{1.19}$ & $59.33_{1.08}$ & $60.16_{0.97}$ \\
\bottomrule
\end{tabular}
\caption{Zero-shot cross-lingual transfer accuracy with standard deviation on XNLI. FT:finetuning; UP: Universal Prompting; \textsc{MV}: multilingual verbalizer. Reported results are averaged with 5 random seeds.}
\label{table:xnli-variance}
\end{table}

%% file: paws-x-variance.tex
\begin{table}[t]
\centering
\scriptsize
\setlength{\tabcolsep}{0.8mm}
\begin{tabular}{c|lcccccccc}
\toprule
 & {\bf Method} & {\bf EN} & {\bf DE} & {\bf ES} & {\bf FR} & {\bf JA} & {\bf KO} & {\bf ZH} & {\bf Avg.} \\
\midrule
\multirow{5}{*}{\textit{256shots}}
& FT & $63.18_{5.69}$& $60.81_{5.04}$ & $60.95_{3.83}$ & $61.39_{4.38}$ & $58.60_{3.85}$ & $58.48_{2.59}$ & $59.78_{4.71}$ & $60.46_{4.23}$ \\
& UP & $65.50_{3.98}$ & $62.21_{3.90}$ & $63.24_{3.52}$ & $62.82_{4.31}$  & $54.11_{5.37}$ & $54.30_{4.98}$ & $55.99_{4.64}$ & $59.74_{4.12}$ \\
& \textsc{Ours} & $\textbf{71.87}_{5.03}$ & $\textbf{68.59}_{4.50}$ & $\textbf{69.10}_{4.92}$ & $\textbf{69.02}_{3.97}$ & $\textbf{60.41}_{2.69}$ & $\textbf{60.88}_{1.76}$ & $\textbf{62.75}_{3.46}$ & $\textbf{66.09}_{3.62}$ \\
& \text{ } \textsc{w/o MV} & $69.06_{4.58}$ & $66.26_{4.13}$ & $66.47_{4.05}$ & $65.79_{4.13}$ & $59.28_{5.36}$ & $58.34_{4.24}$ & $60.77_{6.00}$ & $63.71_{4.37}$ \\
& \text{ } \textsc{w/o Mixup} & $70.95_{4.17}$ & $67.14_{3.53}$ & $67.58_{4.36}$ & $67.63_{4.06}$ & $59.01_{1.67}$ & $60.44_{1.54}$ & $61.16_{2.41}$ & $64.84_{2.91}$ \\
\midrule
\multirow{5}{*}{\textit{512shots}}
& FT & $77.64_{8.38}$ & $73.41_{6.30}$ & $73.19_{6.84}$ & $74.33_{6.47}$ &$65.55_{3.89}$ & $65.19_{3.81}$ & $68.25_{5.28}$ & $71.08_{5.81}$  \\ 
& UP & $83.31_{2.43}$ & $76.18_{1.93}$ & $77.63_{1.73}$ & $77.42_{1.64}$ & $63.41_{2.39}$ & $65.03_{2.12}$ & $68.06_{0.78}$ & $73.01_{1.52}$ \\
& \textsc{Ours} & $\textbf{84.97}_{1.60}$ & $\textbf{78.63}_{1.36}$ & $79.60_{1.31}$ & $\textbf{80.48}_{1.09}$ & $\textbf{67.86}_{1.27}$ & $68.13_{0.88}$ & $\textbf{72.34}_{1.47}$ & $\textbf{76.00}_{1.04}$ \\
& \text{ } \textsc{w/o MV} & $84.81_{1.19}$ & $78.56_{0.78}$ & $\textbf{79.67}_{0.81}$ &$79.64_{0.36}$ & $67.04_{1.85}$ & $\textbf{68.34}_{0.88}$ & $71.50_{1.76}$ & $75.65_{0.64}$ \\
& \text{ } \textsc{w/o Mixup} & $84.84_{1.57}$ & $77.85_{1.94}$ & $79.36_{2.01}$ & $79.69_{1.99}$ & $66.76_{3.24}$ & $68.03_{2.19}$ & $71.03_{2.65}$ & $75.37_{2.00}$ \\
\bottomrule
\end{tabular}%
\caption{Zero-shot cross-lingual transfer accuracy with standard deviation on PAWS-X. FT:finetuning; UP: Universal Prompting; \textsc{MV}: multilingual verbalizer. Reported results are averaged with 5 random seeds.}
\vspace{-0.2cm}
\label{table:pawsx-variance}
\end{table}